\documentclass[preprint,review,12pt]{elsarticle}

\usepackage{amssymb}
\usepackage{verbatim}
\usepackage{paralist,bbding,pifont}
\usepackage{subcaption}
\usepackage{array}
\usepackage{pifont}
\usepackage{soul, color, xcolor}
\usepackage{verbatim}
\usepackage{array}
\usepackage{makecell}
\usepackage{amsmath,amssymb,amsfonts}
\usepackage{graphicx}
\usepackage{textcomp}
\usepackage{xcolor}
\usepackage{booktabs}
\usepackage{bbding}
\usepackage{multirow}

\usepackage{multicol}
\usepackage{amsmath}
\usepackage{amsthm}%
\usepackage{mathrsfs}%
\usepackage{textcomp}%
\usepackage{manyfoot}%
\usepackage{booktabs}%
\usepackage{algorithm}%
\usepackage{algorithmicx}%
\usepackage{algpseudocode}%
\usepackage{listings}%
\usepackage{stfloats}
\usepackage{xcolor}
\usepackage{graphicx}
\usepackage{caption}
\usepackage{ragged2e}
\usepackage{wrapfig}
\usepackage{url} 
\usepackage{hyperref}
\usepackage{cleveref}
\hypersetup{
hidelinks,
colorlinks=true,
linkcolor=red,
}

\usepackage{booktabs}
\usepackage{bbding}

\journal{Pattern Recognition}

\begin{document}

\begin{frontmatter}



\title{CTNeRF: Cross-Time Transformer for Dynamic Neural Radiance Field from Monocular Video}


\author[Durham]{Xingyu~Miao}\ead{xingyu.miao@@durham.ac.uk}
\author[astar]{Yang~Bai}\ead{bai\_yang@ihpc.a-star.edu.sg}
\author[Durham]{Haoran~Duan}\ead{haoran.duan@ieee.org}
\author[Durham]{Fan~Wan}\ead{fan.wan@durham.ac.uk}
\author[Tencent]{Yawen Huang}\ead{yawenhuang@tencent.com}
\author[Durham]{Yang Long\corref{mycorrespondingauthor}}\ead{yang.long@ieee.org}
\author[Tencent]{Yefeng Zheng}\ead{yefengzheng@tencent.com}


\cortext[mycorrespondingauthor]{Corresponding author}

\address[Durham]{Department of Computer Science, Durham University, UK.}
\address[astar]{Institute of High Performance Computing (IHPC), ASTAR, Singapore}
\address[Tencent]{Jarvis Research Center, Tencent YouTu Lab, China.}


\begin{abstract}
The goal of our work is to generate high-quality novel views from monocular videos of complex and dynamic scenes. Prior methods, such as DynamicNeRF, have shown impressive performance by leveraging time-varying dynamic radiation fields. However, these methods have limitations when it comes to accurately modeling the motion of complex objects, which can lead to inaccurate and blurry renderings of details. To address this limitation, we propose a novel approach that builds upon a recent generalization NeRF, which aggregates nearby views onto new viewpoints. However, such methods are typically only effective for static scenes. To overcome this challenge, we introduce a module that operates in both the time and frequency domains to aggregate the features of object motion. This allows us to learn the relationship between frames and generate higher-quality images. Our experiments demonstrate significant improvements over state-of-the-art methods on dynamic scene datasets. Specifically, our approach outperforms existing methods in terms of both the accuracy and visual quality of the synthesized views. Our code is available on \url{https://github.com/xingy038/CTNeRF}.

\end{abstract}


\end{frontmatter}


\section{Introduction}
Realistically rendering and presenting dynamic real-world scenes is a highly challenging research topic, with diverse applications in fields such as film production and virtual reality \cite{miller2005interactive,collet2015high,smolic20063d}. However, accurately modeling these scenes using traditional mesh-based methods can be difficult due to the complex movements of multiple objects and changes in factors like mirroring and transparency that occur during these movements. While multi-view-based methods have shown better results, they come with their own limitations. These methods require a large number of cameras, resulting in high costs and technical challenges like synchronization and data processing \cite{carranza2003free,zitnick2004high,orts2016holoportation,broxton2020immersive}. Additionally, they are not easily applicable in daily life scenarios. Although reconstruction from monocular videos is a promising approach for scene reconstruction, novel view synthesis for monocular videos of dynamic scenes is more challenging.

Recent advancements in deep learning have made significant breakthroughs in novel view synthesis, with Neural Radiative Fields (NeRF) \cite{mildenhall2022nerf,xian2021space} being one of the most notable contributions to this area. NeRF employs the position and viewing direction of a given image as a query, and employs volume rendering to generate the color of each pixel. However, these methods are primarily designed for static scenes and do not perform optimally when dealing with dynamic objects or scenes. To address this limitation, recent research has explored the application of this approach to monocular dynamic video \cite{du2021neural,gao2021dynamic}. For example, some studies have focused on learning a deformable warp field \cite{park2021hypernerf} or a neural scene flow between adjacent frames \cite{li2021neural,li2022dynibar,gao2021dynamic,miao2023ds}. These efforts aim to extend the utility of NeRF and enable more robust and accurate synthesis of dynamic scenes. Despite the success achieved by NeRF-based methods for dynamic scenes, they still have some limitations. For instance, deformable warp field methods such as Nerfies \cite{park2021nerfies} can handle long sequences but may not perform well for dynamic scenes with complex object motion. On the other hand, neural scene flow or neural trajectory methods like NSFF \cite{li2021neural} can handle large movements in dynamic scenes, but their effectiveness is highly dependent on the accuracy of the predicted scene flow or trajectory.

We propose a novel approach that can be applied to dynamic scenes, enabling the handling of more complex motions and improving the rendering results. Our method draws inspiration from recent research on rendering static scenes \cite{wang2021ibrnet,wang2022attention,chen2021mvsnerf,liu2022neural}, where local image features are synthesized by aggregating them along epipolar lines from nearby views to enhance the rendering process. However, the apparent limitations assumed by these methods are violated by scenes in motion, making them unsuitable for direct application to dynamic scenes. To overcome this challenge, we have designed a module that aggregates changes in ray due to motion in the ray space, along with the obtained multi-view features \cite{yang2024geometric}. This enables us to accurately consider both temporal and spatial changes in geometry and appearance, resulting in better rendering of dynamic scenes. More specifically, we first input the extracted feature vector into a cross-time transformer. Next, we input the feature aggregated with time information into a ray transformer to find the relationship between the sampling points on the ray and obtain the aggregated feature. In addition, to strengthen the spatial-temporal relationship of feature vectors, we use a 2D fast Fourier transform frequency-domain feature aggregation module to obtain the aggregated features. Finally, we feed the fused feature vectors of these two features together with the queried rays into residual-based MLPs to output color and density. Experimental results demonstrate that our method can synthesize new views with high quality. Furthermore, compared to previous methods, our approach can render higher-quality ground-truth details of ground truth in dynamic regions. In summary, the contributions of our work are as follows:

\begin{itemize}
  \item[1)]
  A novel dynamic neural rendering field for dynamic monocular video, which can aggregate multi-view feature vectors to improve rendering novel view quality.
  \item[2)]
  The aggregation of multi-frame feature vectors may lead to the potential loss or merging of intricate details into other features, thereby compromising the retention of crucial characteristics from the original data. To address this issue, we introduce a Ray-based cross-time transformer.
  \item[3)]
  To mitigate potential blurring during feature aggregation, we introduce a Global Spatio-Temporal Filter.
  \item[4)]
  Extensive experiments show that our method achieves superior novel view synthesis of dynamic scenes.
\end{itemize}

\section{Related Work}
\subsection{Novel view synthesis} 
Recently, neural implicit representation methods like NeRF \cite{liu2020neural,mildenhall2022nerf,xian2021space,mildenhall2021nerf,xiangli2022bungeenerf,xu2022point,yu2021pixelnerf} have demonstrated significant potential for achieving high-quality rendering. NeRF employs multi-layer perceptrons (MLPs) to implicitly represent continuous scenes, yielding impressive view synthesis results. Despite their progress, NeRF-based methods necessitate training separate models for each scene, with optimization demanding varying training times. Applying these methods faces some other challenges, including unknown camera poses, boundary blur, and observation noise. For unknown camera poses, Li et al. \cite{10197499} proposed a novel online scene representation method that can simultaneously learn to represent the target scene and estimate the camera pose from the RGB-D stream. For boundary-blurring, Barron et al. proposed Mip-NeRF \cite{barron2021mip}, which uses sampling of cones instead of rays and considers scale information by integrating position encoding, so that the scene is represented in a scale of continuous values, and the rendering result is anti-aliased. In addition, variations of NeRF-based methods, such as PixelNeRF \cite{yu2021pixelnerf}, MVSNeRF \cite{chen2021mvsnerf}, and IBRNet \cite{wang2021ibrnet}, exhibit promise in incorporating feature information to generalize to unseen scenes. However, their primary focus is on static scenarios, neglecting dynamic scenes with objects or cameras in motion. These methods estimate a 3D representation of a scene using multiple input images, which they then leverage for rendering novel views. Nevertheless, their applicability is limited in dynamic scenes, where assumptions of scene stability may lead to inaccuracies or artifacts in rendered images. Our work addresses this limitation by extending the approach to more challenging dynamic scenarios, concentrating on modeling complex object motion and synthesizing higher-quality novel views.

\subsection{Dynamic region view synthesis} 
With NeRF \cite{liu2020neural,mildenhall2022nerf,xian2021space} demonstrating impressive results in view synthesis tasks, several works have attempted to extend NeRF to tackle dynamic new view synthesis challenges \cite{gao2021dynamic,li2022dynibar,li2021neural, tretschk2021non}. These methods can be classified into two main directions. The first direction involves using deformation fields to represent scenes \cite{tretschk2021non,park2021hypernerf,park2021nerfies,pumarola2021d}. While this approach can handle long sequences of videos, its primary challenge is dealing with large motions in the scene. As this method typically warps the scene from the same frame, it can result in a lack of continuity throughout the entire sequence, such as Nerfies \cite{park2021nerfies} and HyperNeRF \cite{park2021hypernerf}. The second approach is based on the time-varying 4D radiance fields approach \cite{gao2022monocular,li2021neural,wang2021neural}. These methods model dynamic scenes as time-varying continuous functions of appearance, geometry, and 3D scene motion by predicting the scene flow field. Although such methods can capture fast and complex motion in the scene, they usually require more accurate scene flow or trajectory guidance and cannot handle non-rigid deformation well. Our proposed method aggregates feature from nearby views to effectively handle this situation and improve rendering results.

\begin{figure*}
  \centering
  \includegraphics[width=0.8\linewidth]{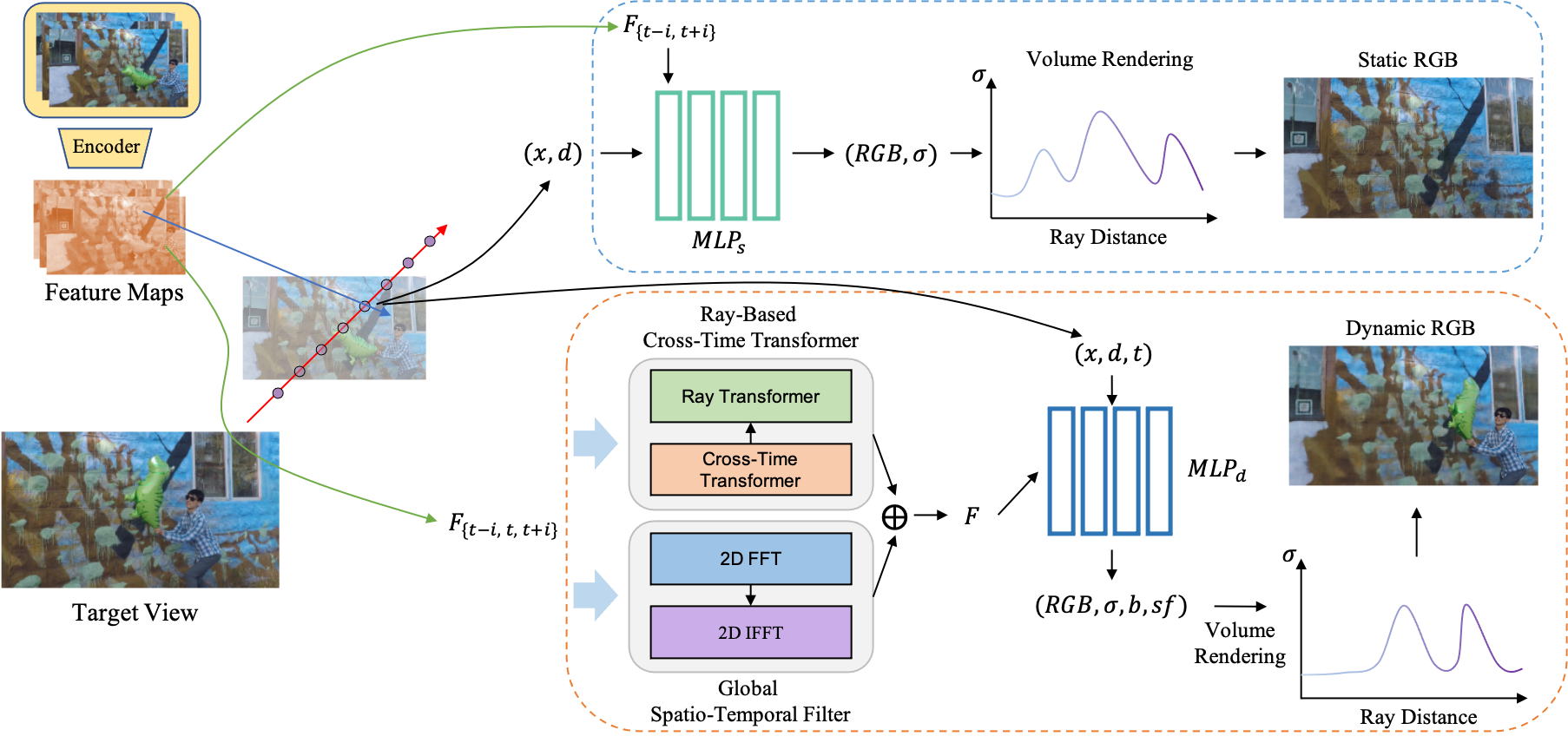}
  \caption{The pipeline of our model. Our model is composed of two main parts, each responsible for handling a different aspect of the input data. One component focuses on the static background, while the other deals with the dynamic foreground. These two sets of values are then blended together to obtain the final novel view.}
  \label{fig:2}
\end{figure*}
\section{Proposed Method}
In this section, we present the proposed methods with the goal of enabling the trained model to query new viewpoints at any time and angle within a monocular video of a dynamic scene. Our system pipeline (\Cref{fig:2}) can be divided into two parts: one part focuses on the static background, while the other part handles the dynamic foreground, and finally blends the two through blending to obtain the reconstructed video. Furthermore, similar to other time-varying NeRF-based techniques, we first optimize the model to reconstruct the input frame, before being utilized for rendering novel views.  Instead of directly encoding 3D color and density in the weights of the MLPs like recent dynamic NeRF methods \cite{gao2021dynamic, li2021neural}, we borrow the idea of a recent generalized NeRF to aggregate features from views near the target view to enhance rendering. Below we describe our approach to multi-feature aggregation, a ray-based cross-time aggregation module, and a boost module via frequency-domain effects.

\subsection{Multi-view aggregation}
We leverage two models to reconstruct the static and the dynamic area respectively and finally blend the color and density of the two through a blending value predicted by the dynamic model in the interval $[0,1]$ to obtain the final reconstructed image.
\subsubsection{Static Region Feature Aggregation}
For the static region, we simply adopt the projection methods to query the position of the camera ray projected on the image coordinates at a certain point in the space. And then the corresponding feature vectors can be obtained by the method of bilinear difference. More specifically, the feature vectors of two adjacent frames are queried using the camera ray from the target viewpoint, which can be expressed as:
\begin{equation}
x_{t\pm i} = P_{\pm i}x_t \in \mathbb{R}^3
\label{eq:1}
\end{equation}
where $x_t$ is a point in space on the camera ray on the target view, the $x_{t\pm i}\in \mathbb{R}^3$ is the adjacent view, and the $P_{\pm i}=[R_{\pm i},T_{\pm i}]\in \mathbb{R}^{4\times4}$ is the camera parameters, note that we set $ i = 1$. And the queried feature can be expressed as:
\begin{equation}
F_{t\pm i}=E(proj\left \langle x_{t\pm i} \right \rangle)\in \mathbb{R}^d
\label{eq:2}
\end{equation}
$proj\left \langle \cdot \right \rangle$ represents the coordinates of the point projection image in space, and then uses a feature extractor $E(\cdot)$ to extract the features of the image, and finally obtains the query feature vector $F_{t\pm i}\in \mathbb{R}^d$. The RGB $c$ and density $\sigma$ of the static region can use an MLP to query, which can be expressed as:
\begin{equation}
MLP_{\theta s}(x,y,z,t,d_t,F_{t\pm i}) = (c_s,\sigma_s)
\label{eq:3}
\end{equation}
the inputs include extracted feature vectors $F_{t\pm i}\in \mathbb{R}^d$, target view direction $d_t\in \mathbb{R}^3$, and space coordinates $x,y,z$.

\begin{figure}
  \centering
  \includegraphics[width=0.7\linewidth]{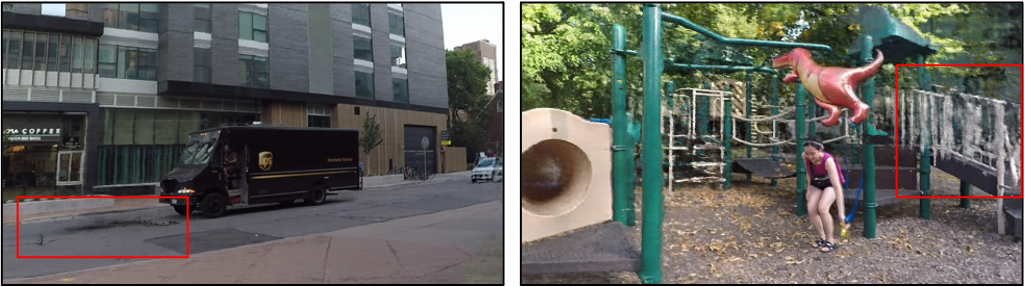}
  \caption{Aggregating feature vectors in an epipolar-aligned manner will cause errors in the rendering of the model, resulting in artifacts that degrade the quality of the model rendering novel views.}
  \label{fig:artifacts}
\end{figure}

\subsubsection{Dynamic Region Feature Aggregation}
For the dynamic region, we cannot use the same method as the static region to aggregate features. The object movement violates the static hypothesis, computing adjacent frames only with camera parameters cannot handle this change. Therefore, inspired by recent neural scene flow work \cite{gao2021dynamic, li2021neural}, we first use predicted scene flow to warp the camera rays to describe the motion of a point in space in the scene, which can be expressed as:
\begin{equation}
x_{t+1}=x_t+s_{fw}\in \mathbb{R}^3
\label{eq:4}
\end{equation}
\begin{equation}
x_{t-1}=x_t+s_{bw}\in \mathbb{R}^3
\label{eq:5}
\end{equation}
where, $s_{fw}\in \mathbb{R}^3$ and $s_{bw}\in \mathbb{R}^3$ are the predicted scene flow. And then, we can obtain the corresponding feature vectors using \Cref{eq:2}. Note that we use a four-layer MLP to predict this scene flow, while also outputting a blend value to weigh the color and density of static and dynamic regions. Thus, it can be expressed as:

\begin{equation}
MLP_{\theta d}(x,y,z,t,d_t,F_{t\pm i}) = (c_d,\sigma_d,s_{fw},s_{bw},b)
\label{eq:6}
\end{equation}
$s_{fw}\in \mathbb{R}^3$ and $s_{bw}\in \mathbb{R}^3$ are the predicted scene flow, $b$ is the predicted blending value and the $i = {\{0, 1\}}$.
\subsubsection{Combining static and dynamic models}
Time-varying-based dynamic models typically undergo very much deformation to reliably infer correspondences over larger time intervals, however, static regions should be consistent. In order to render complete and high-quality content in the static area of the new view composition, we follow the idea of NSFF \cite{li2021neural} and use two separate models (static and dynamic) to model the entire scene. Through the above methods, we can obtain static and dynamic colors $c_{s},c_{d}$ and densities $\sigma_{s},\sigma_{d}$ respectively, then the volumetric radiance field can then be rendered into a 2D image via:
\begin{equation}
\begin{split}
C_{full}(r)&=\int_{t_n}^{t_f} (T_d(t)\sigma_d(t)c_d(t)b\\
&+T_s(t)\sigma_s(t)c_s(t)(1-b)) dt
\end{split}
\label{eq:7}
\end{equation}
\begin{equation}
T_{\{s,d\}}(t)=exp\left(-\int_{t_n}^{t}\sigma_{\{s,d\}}(s)ds\right)
\label{eq:8}
\end{equation}

The rendered pixel values for camera ray $r$ can then be compared to the corresponding ground truth pixel values:
\begin{equation}
\mathcal{L}_{pho}=\sum_{r}\left \| \hat C(r)- C_{gt}(r)\right \|_2^2
\label{eq:9}
\end{equation}

where $C(r)$ includes the static, dynamic, and blended regions. Directly aggregating these features can enhance the representation of target feature maps, and the effect of improving quality can be obtained in the reconstruction stage. However, through our observations during the rendering process of the novel view, artifacts of adjacent frames (see \Cref{fig:artifacts}) manifest in the novel view, significantly compromising the performance of our model. In addition, we observed that our methods will produce some non-rigid deforms when rendering novel views of dynamic scenes, it also will affect the quality of novel views synthesis. Thus, we propose a ray-based cross-time (RBCT) aggregation module to handle this issue.

\begin{figure*}
  \centering
  \includegraphics[width=0.8\linewidth]{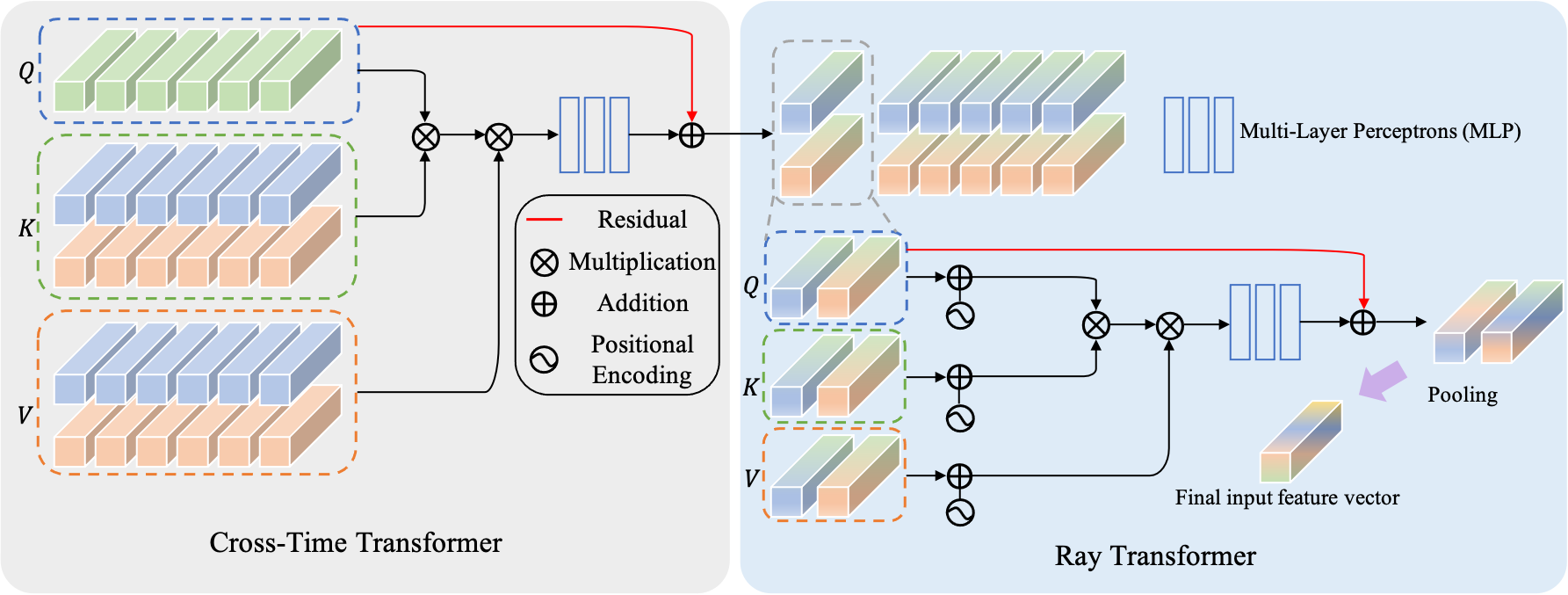}
  \caption{The pipeline of the RBCT module. The model consists of two main components: the cross-time transformer on the left and the ray transformer on the right. The left component takes a set of feature vectors from consecutive frames as input and applies cross-time attention to aggregate these vectors with the current frame. The resulting feature vector is then passed to the right component, which uses ray attention to aggregate feature vectors from multiple sampling points along each ray. Finally, a pooling operation is applied to these vectors to obtain the final aggregated feature vector.}
  \label{fig:3}
\end{figure*}

\subsection{Ray-Based Cross-Time Aggregation Module}
\subsubsection{Cross-Time Aggregation}
By means of the aforementioned method, the correlation between the frontal and posterior frames can be integrated using the camera rays from the current viewing angle. However, it should be noted that there exists a temporal relationship among the input feature vectors of adjacent frames, and a direct aggregation would overlook this temporal variation. Therefore, we design a cross-time converter to enable the feature vector of the target frame to attentively focus on the variations in the feature vector of its adjacent frame (as illustrated in the left figure, see \Cref{fig:3}). The cross-time transformer leverages a classic cross-attention mechanism \cite{vaswani2017attention,duan2023dynamic,wu2024spatial, cao2024mhsan} that allows for a variable number of inputs, given the frame-to-frame changes involved:
\begin{equation}
\hat F_{t\pm i}=\text{Cross-Time\ Transformer}(F_{t\pm i}, F_t)
\label{eq:10}
\end{equation}
For a given target frame \(F_t\), the query vector \(Q_t\) is derived by linearly transforming the original feature vector:
\begin{equation}
Q_t = W_Q \cdot F_t
\end{equation}
Similarly, key (\(K_{t\pm i}\)) and value (\(V_{t\pm i}\)) vectors for adjacent frames (\(F_{t\pm i}\)) are obtained through linear transformations:
\begin{equation}
K_{t\pm i} = W_K \cdot F_{t\pm i}, \quad V_{t\pm i} = W_V \cdot F_{t\pm i}
\end{equation}
The attention scores are computed using the dot product of the query and key vectors, scaled by a factor \(\sqrt{d_k}\):
\begin{equation}
Attn = \frac{Q_t \cdot K_{t\pm i}^T}{\sqrt{d_k}}
\end{equation}
Finally, the target frame's feature vector \(\hat F_{t\pm i}\) is updated by computing a weighted sum of the value vectors:
\begin{equation}
\hat F_{t\pm i} = Softmax(Attn) \cdot V_{t\pm i}
\end{equation}
Our ablation studies demonstrate that this approach effectively enhances the quality of synthesized images.

\subsubsection{Ray Aggregation}
While aggregating camera views, we have successfully explored the relationship between the camera ray's point in space and its corresponding feature vector. However, one crucial aspect that has been overlooked is the relationship between feature vectors from adjacent frames. To address this limitation and improve the overall representation, we introduced a cross-time transformer mechanism, allowing the target frame to focus on the inter-frame relationships and enhancing the global correlation. Despite the progress achieved with the cross-time transformer, we encountered an issue. Specifically, this approach failed to accurately associate the local relationship between the camera ray's sampling point and its corresponding per-frame feature vector.  To overcome this limitation, we propose a method similar to the depth bin utilized in stereo-matching algorithms when calculating the cost volume \cite{yao2018mvsnet}. This involves considering each sampling point along the entire ray to match a specific pixel in the form of a matching score.

More specifically, we introduce a new ray transformer that enables the mutual focus of feature vectors corresponding to samples on a ray. The ray transformer is composed of two core components of the classical transformer \cite{vaswani2017attention}: position encoding and self-attention. It can be expressed as:
\begin{equation}
\mathcal{F}=\text{Ray\ Transformer}(\hat F_{t\pm i})
\label{eq:11}
\end{equation}
Specifically, we define:
\begin{equation}
Q = W_Q \cdot \hat F_{t\pm i}, K = W_K \cdot \hat F_{t\pm i}, V = W_V \cdot \hat F_{t\pm i}
\end{equation}
Here, \(W_Q\), \(W_K\), and \(W_V\) are learned weight matrices. Subsequently, we compute attention scores and utilize these attention weights to calculate a weighted average, yielding a new feature representation:
\begin{equation}
\mathcal{F} = Softmax(\frac{Q \cdot K^T}{\sqrt{d_k}}) \cdot V
\end{equation}
Given $M$ (we set $M$ to 64) samples along a ray, our ray transformer transforms the output of its input cross-time transformer, resulting in an aggregated feature vector. The introduction of this ray transformer allows the model to focus on the feature vectors corresponding to samples along the ray in adjacent frames, capturing finer local relationships and addressing the limitations of the previous cross-time transformer in this regard. Our ablation experiments demonstrate that the proposed ray transformer significantly improves the quality of the final synthesized image.

\subsection{Frequency Domain Aggregation Module}
We introduce a novel spatio-temporal feature learning module termed the Global Spatio-Temporal Filter (GSTF), inspired by recent advancements in frequency-domain-based methodologies \cite{zhou2024frequency} aimed at enhancing the rendering quality of new views \cite{rao2021global}. The primary objective of GSTF is to elevate the representation of feature vectors by capturing both spatial and temporal relationships through specialized frequency filters. In our approach, GSTF is meticulously crafted to discern and learn distinct frequency filters at each spatial location. This enables the modeling of temporal variations within feature vectors across different spatial positions. The core mechanism involves the transformation of both temporal and spatial features at each location into frequency feature spectra. This transformation is achieved through a two-dimensional Fast Fourier Transform (FFT) \cite{cooley1965algorithm}. The frequency filter, learned through GSTF, acts as a modulator on this transformed spectrum. Subsequently, we revert this modulated spectrum back to the time domain using an inverse FFT. This comprehensive process allows GSTF to effectively encode the intricate interplay between time and space in the feature vectors, contributing to the improvement of new view rendering. To gain a better understanding of our GSTF design, let's first review the convolution theorem in the field of digital signal processing \cite{oppenheim1997signals}. Given a sequence of feature signals with $T$ points ($f[t], 0\le t\le T-1$), we can calculate its discrete spectrum $S[k]$ using Discrete Fourier Transform (DFT) via:
\begin{equation}
S[k]=\sum_{t=0}^{T-1} f[t] e^{-j(2 \pi / T) k t}, 0 \leq k \leq T-1
\label{eq:12}
\end{equation}

In the equation above, $j$ represents the imaginary unit. The Discrete Fourier Transform (DFT) is a one-to-one orthogonality decomposition. Moreover, we can use the DFT outputs to reconstruct input signals using Inverse Discrete Fourier Transform (IDFT) via:
\begin{equation}
f[t]=\frac{1}{T} \sum_{k=0}^{T-1} S[k] e^{j(2 \pi / T) k t}, 0 \leq t \leq T-1
\label{eq:13}
\end{equation}
Specifically, we first convert the features into frequency domain signals. Next, these frequency domain signals are filtered, and then the filtered signals are reconstructed into time domain features. Our GSTF can be easily used in modern deep-learning frameworks such as PyTorch \cite{paszke2019pytorch}, The pseudocode of PyTorch is shown in \Cref{alg:1}. Finally, these time-domain features are merged with the RBCT-processed features to achieve effective aggregation of time-domain and frequency-domain features. It can be expressed as:
\begin{equation}
F = f[t] + \mathcal{F}
\end{equation}
Where $f[t]$ is the frequency-domain features, $\mathcal{F}$ is the time-domain features, and $F$ is the aggregated features. Through ablation experiments, we demonstrate that our global filtering mechanism is an effective spatial information mixing method. The GSTF module improves the texture quality of new views and alleviates blurring and artifacts often observed in new view synthesis. 

\begin{algorithm}[t]
  \caption{Global Spatio-Temporal Filter}
  \label{alg:1}
  \begin{algorithmic}[1]
    \State Initialization: learnable weight w
    \State x = torch.fft.rfft2(x, dim=(0,1))
    \State x = x * w
    \State x = torch.fft.irfft2(x, dim=(0,1))
  \end{algorithmic}
\end{algorithm}

\subsection{Regularization}
It is well known that monocular video reconstruction of complex dynamic scenes is an ill-posed problem, and using only a photometric error for supervision cannot avoid local minima. Therefore, many regularization strategies have been used in previous work \cite{gao2021dynamic,li2021neural}. We continue to use the previous strategy and add several regularization items. Specifically, it includes the following three parts:
\begin{equation}
\mathcal{L} =\mathcal{L}_{data}+\mathcal{L}_{small}+\mathcal{L}_{pho}
\label{eq:14}
\end{equation}
The $\mathcal{L}_{data}$ is a data-driven prior regularization term, composed of pre-trained monocular depth estimation network and optical flow estimation network consistency prior \cite{gao2021dynamic}. The $\mathcal{L}_{pho}$ is provided by \Cref{eq:9}. Our model is highly dependent on the accuracy of scene flow, thus we provide an additional regularization term for scene flow. The $\mathcal{L}_{small}=\left \| s_{fw} \right \| _1 +\left \| s_{bw} \right \| _1+\left \| s_{fw}+s_{bw} \right \| _1$ is a regularization term that minimizes scene flow.

\subsection{Architecture}
\begin{figure}
  \centering
  \includegraphics[width=0.5\linewidth]{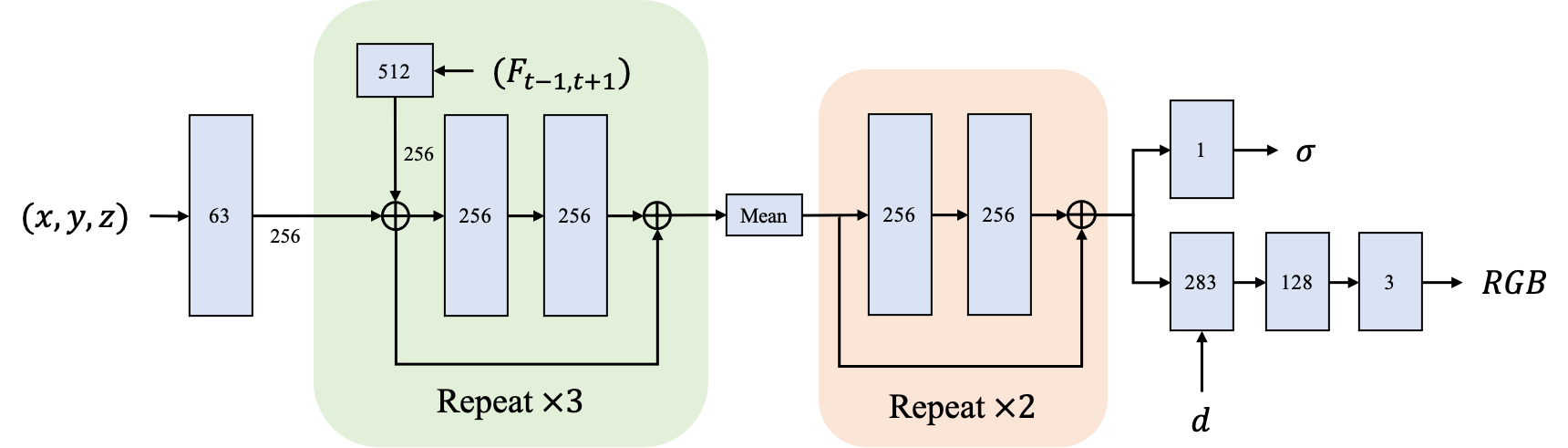}\hfill
  \includegraphics[width=0.5\linewidth]{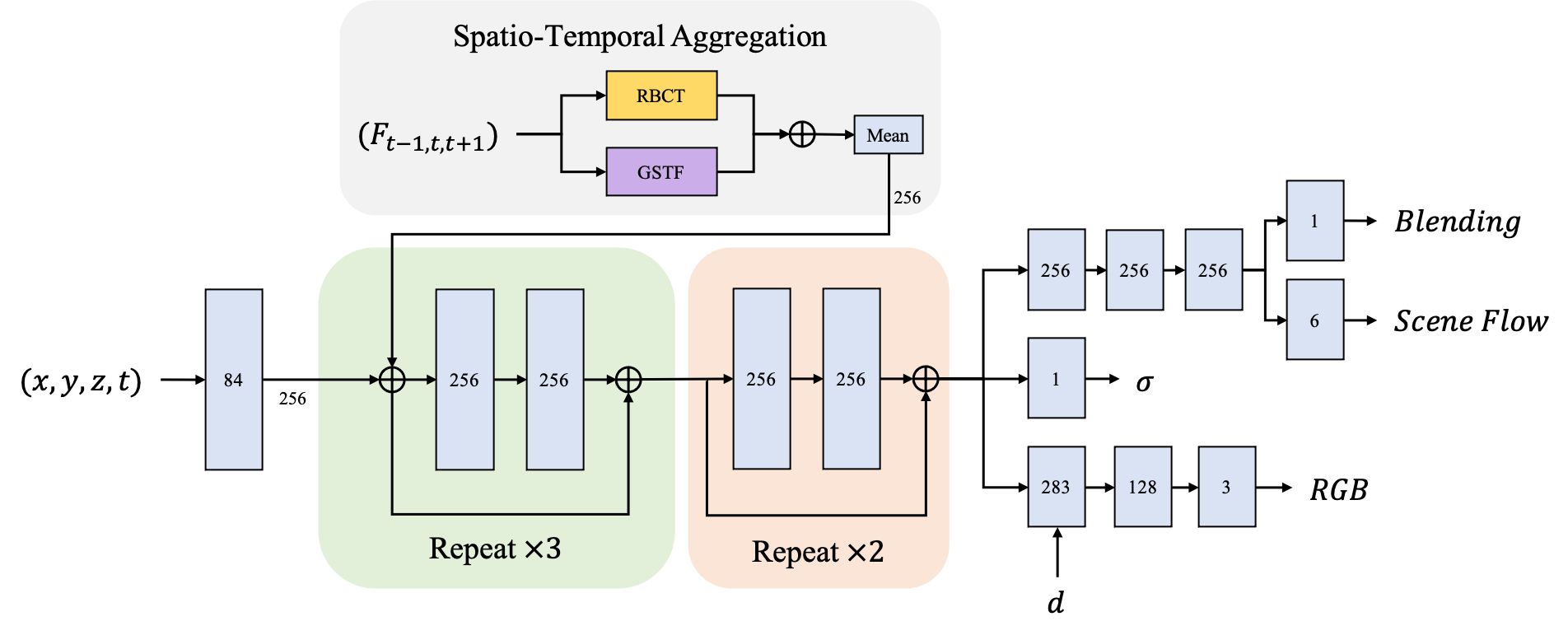}
  \caption{Network architectures of our static and dynamic representations.}
  \label{fig:both}
\end{figure}

Our approach utilizes two distinct types of models: static and dynamic. The architecture of static and dynamic model is depicted in \Cref{fig:both}.
\section{Experiment}
\subsection{Implementation details}
Our model uses ResNet34 \cite{He_2016_CVPR} as the encoder to extract feature maps. There are some differences between the static and dynamic models. For the static model, we use ResNet-based MLPs block, while for the dynamic model, we add four additional layers of MLPs to predict scene flow and mixed values. We found that the ResNet-based MLP block is difficult to train for accurate scene flow. More details about our model can be found in the supplementary material. We first train the static model for 300K steps and then fix it to train the dynamic model for 200K steps. We use frames $t-1, t$, and $t+1$ as input to extract feature vectors. Note that we only choose $t-1,t$ and $t,t+1$ when selecting the first and last frames as input.
\begin{figure*}
  \centering
  \includegraphics[width=\linewidth]{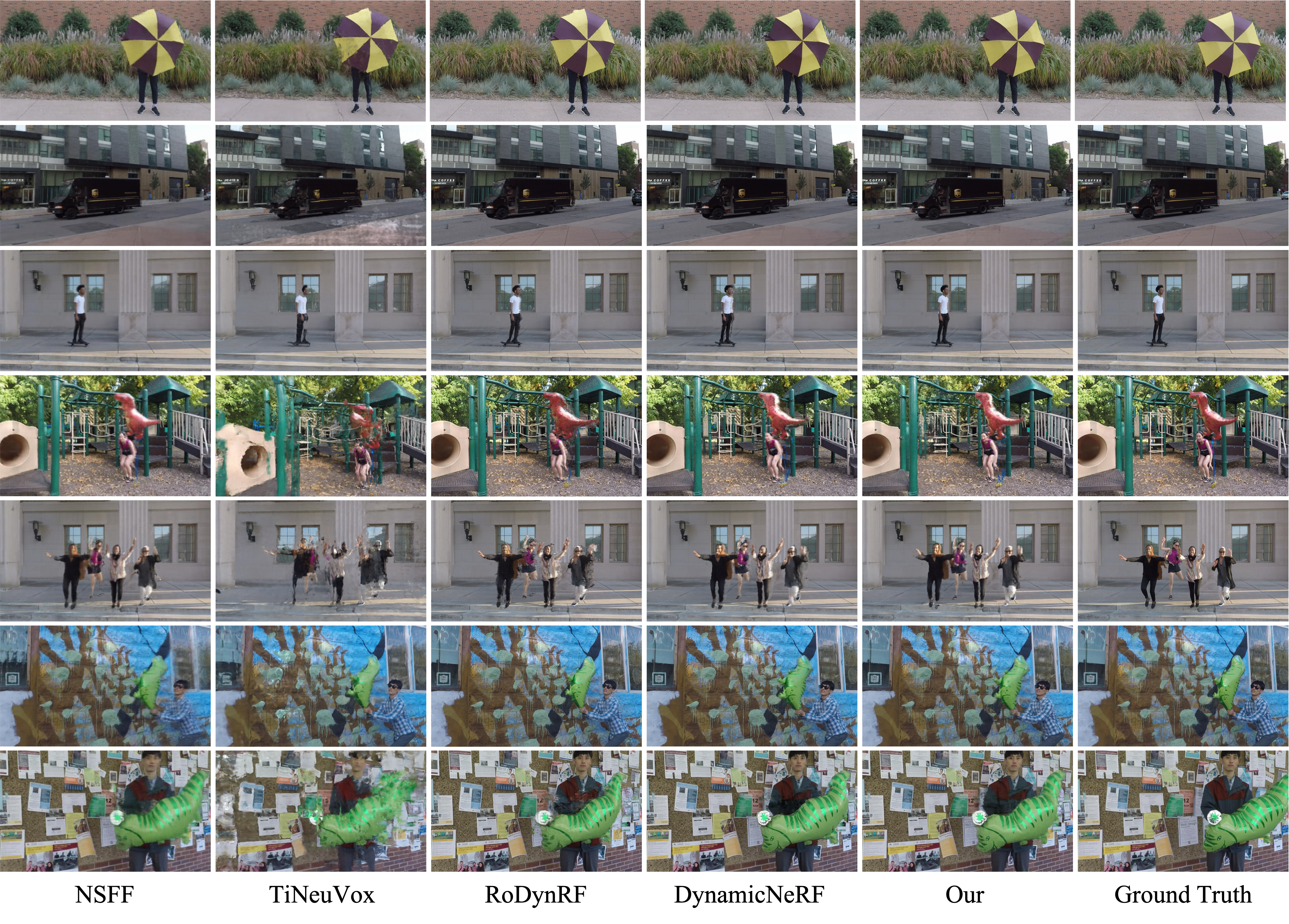}
  \caption{Novel view synthetic qualitative results on Nvidia Dynamic Scene Dataset \cite{yoon2020novel}. In contrast to other NeRF-based approaches, our outcomes exhibit enhanced clarity, capturing finer details that closely approximate ground truth, particularly in dynamic regions.}
  \label{fig:4}
\end{figure*}
\subsection{Model Parameters and Inference Time}
\begin{wraptable}{r}{0.5\textwidth}
    \centering
    \resizebox{\linewidth}{!}{
    \begin{tabular}{cccc}
    \toprule
    Method      & Parameters & Inference speed & Training Time   \\ 
    \midrule
    DynamicNeRF & 4.59M      & 8 s/Frame       & about 21 hours \\
    Our         & 12.21M     & 10 s/Frame      & about 24 hours  \\ 
    \bottomrule
    \end{tabular}
    }
    \caption{Parameters, inference speed, and training time.}
    \label{tab:1}
\end{wraptable}
We present our model's Parameters, inference speed, and training time. The amount of our speed parameters has increased compared to our baseline, but in more complex scenes, our baseline cannot correctly obtain dynamic scenes (As shown in \Cref{fig:quality1}).

\subsection{Dataset}
Our method is assessed on the Nvidia Dynamic Scene Dataset \cite{yoon2020novel}, DAVIS Dataset \cite{perazzi2016benchmark}, and iPhone dataset \cite{gao2022monocular}.

Nvidia Dynamic Scene Dataset comprises nine video sequences captured using a static camera rig of 12 cameras. All cameras capture images simultaneously at 12 different time steps $\{t_0,t_1,\dots,t_{11}\}$, and we obtain a twelve-frame monocular video $\{I_0, I_1,\dots, I_{11}\}$ by sampling the image taken by the $i-th$ camera at time $t_i$. It is worth mentioning that we use a different camera for each video frame to simulate camera motion. The video frames consist of a background that remains stationary throughout the video and a dynamic object that changes over time. We adopt COLMAP \cite{schonberger2016structure} similar to NeRF \cite{mildenhall2021nerf}, to estimate the camera poses, near and far boundaries of the scene, and assume that all cameras share the same intrinsic parameter. We exclude the DynamicFace sequence from our evaluation since COLMAP fails to estimate camera poses for this sequence. Lastly, we resize all video sequences to a resolution of 480 $\times$ 270. The DAVIS Dataset \cite{perazzi2016benchmark} consists of fifty sequences featuring dynamic moving objects, like animals and cars. However, due to limitations in camera movement, COLMAP could only estimate camera poses for six out of the fifty sequences, all of which include ground truth object masks. Finally, for the iPhone Dataset \cite{gao2022monocular}, we conducted a quantitative evaluation using seven dynamic scenes, each accompanied by ground truth images of novel views.

\subsection{Comparison with State-of-the-Art Methods}
\subsubsection{Qualitative Results} 
\begin{figure}[!htb]
  \centering
  \begin{minipage}{0.5\textwidth}
    \includegraphics[width=\linewidth]{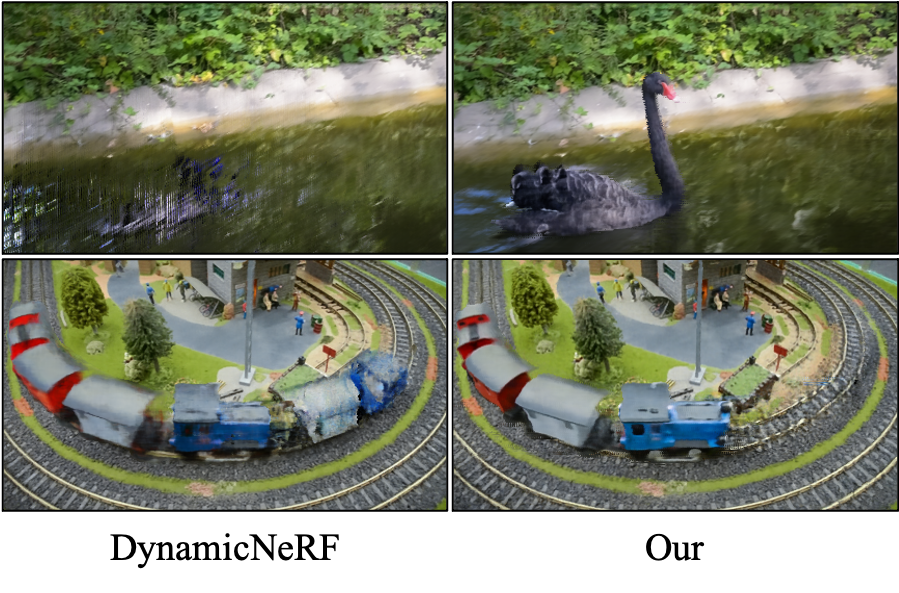}
  \end{minipage}%
    \hspace{1cm}
  \begin{minipage}{0.4\textwidth}
  \caption{Novel view synthetic qualitative results on DAVIS Dataset \cite{perazzi2016benchmark}. Compared to our baseline, our method obtains sharper results and fewer artifacts.}
  \label{fig:quality1}
  \end{minipage}
\end{figure}

We present some visual comparisons on the Nvidia Dynamic Scene Dataset \cite{yoon2020novel} in \Cref{fig:4} and the DAVIS dataset \cite{perazzi2016benchmark} in \Cref{fig:quality1}. The camera poses of most sequences in the DAVIS dataset cannot be estimated by COLMAP. By aggregating the features of adjacent frames, our method generates frames with fewer visual artifacts and obtains results that are closer to ground truth. In contrast, our method exploits feature associations across frames, which yields better visual quality results.

\begin{table*}
  \caption{Novel view synthesis quantitative results on Nvidia Dynamic Scene Dataset \cite{yoon2020novel}. We report average PSNR and LPIPS \cite{zhang2018unreasonable} results by comparison with existing methods. The best performance is bold and the next best is underline.}
  \label{tab:sota}
  \resizebox{\linewidth}{!}{
  \begin{tabular}{lccccccc|c}
    \toprule
\multicolumn{1}{c}{PSNR$\uparrow$ /LPIPS$\downarrow$} & \multicolumn{1}{c}{Jumping} & \multicolumn{1}{c}{Skating} & \multicolumn{1}{c}{Truck} & \multicolumn{1}{c}{Umbrella} & Balloon1      & Balloon2      & Playground    & Average       \\ \hline
NeRF \cite{mildenhall2021nerf}                              & 20.58 / 0.305               & 23.05 / 0.316               & 22.61 / 0.225             & 21.08 / 0.441                & 19.07 / 0.214 & 24.08 / 0.098 & 20.86 / 0.164 & 21.62 / 0.252 \\
NeRF \cite{mildenhall2021nerf} + time                       & 16.72 / 0.489               & 19.23 / 0.542               & 17.17 / 0.403             & 17.17 / 0.752                & 17.33 / 0.304 & 19.67 / 0.236 & 13.80 / 0.444 & 17.30 / 0.453 \\
D-NeRF \cite{pumarola2021d}                       & 22.36 / 0.193              &  22.48 / 0.323               & 24.10 / 0.145             & 21.47 / 0.264                & 19.06 / 0.259 & 20.76 / 0.277 & 20.18 / 0.164 & 21.48 / 0.232 \\
HyperNeRF \cite{park2021hypernerf}                       & 18.34 / 0.302                     &   21.97 / 0.183               & 20.61 / 0.205             &  18.59 / 0.443                &  13.96 / 0.530 & 16.57 / 0.411 &  13.17 / 0.495 & 17.60 / 0.367 \\
TiNeuVox \cite{fang2022fast}                       & 20.81 / 0.247                     &    23.32 / 0.152               &  23.86 / 0.173             &  20.00 / 0.355                &  17.30 / 0.353 & 19.06 / 0.279 &   13.84 / 0.437 & 19.74 / 0.285 \\
Yoon et al. \cite{yoon2020novel}                       & 20.16 / 0.148               & 21.75 / 0.135               & 23.93 / 0.109             & 20.35 / 0.179                & 18.76 / 0.178 & 19.89 / 0.138 & 15.09 / 0.183 & 19.99 / 0.153 \\
Tretschk et al. \cite{tretschk2021non}                  & 19.38 / 0.295               & 23.29 / 0.234               & 19.02 / 0.453             & 19.26 / 0.427                & 16.98 / 0.353 & 22.23 / 0.212 & 14.24 / 0.336 & 19.20 / 0.330 \\
NSFF \cite{li2021neural}                              & 24.12 / 0.146               & \underline{28.91} / 0.135               & 25.94 / 0.171             & 22.58 / 0.302                & 21.40 / 0.225 & 24.09 / 0.228 & 20.91 / 0.220 & 23.99 / 0.205 \\
RoDynRF \cite{liu2023robust}                      & 24.27 / 0.100               &  28.71 / \underline{0.046}              &  \textbf{28.85} / \textbf{0.066}             & \underline{23.25} / \textbf{0.104}                &  \underline{21.81} / \underline{0.122} &  25.58 / 0.064 &  \textbf{25.20} / \textbf{0.052} & \underline{25.38} / \textbf{0.079} \\
DynamicNeRF \cite{gao2021dynamic}                      & \underline{24.61} / \underline{0.144}               & 28.90 / 0.124              & 25.78 / 0.134             & 23.15 / 0.146                & 21.47 / 0.125 & \underline{25.97} / \underline{0.059} & 23.65 / 0.093 & 24.74 / 0.118 \\
 \hline
Our                               &       \textbf{24.35} / \textbf{0.094}                      &        \textbf{33.51} / \textbf{0.034}                     &      \underline{28.27} / \underline{0.084}                     &     \textbf{23.48} / \underline{0.129}                         &     \textbf{22.19} / \textbf{0.111}         &    \textbf{26.86} / \textbf{0.048}          &       \underline{24.28} / \underline{0.077}        &       \textbf{26.17} / \underline{0.082}        \\ 

  \bottomrule
\end{tabular}
}
\end{table*}
\subsubsection{Quantitative Results}
\Cref{tab:sota} presents the quantitative results obtained from the Nvidia Dynamic Scene Dataset \cite{yoon2020novel}. We adopted the evaluation methodology from DynamicNeRF \cite{gao2021dynamic} to synthesize views using the first camera while varying the time on the Nvidia Dynamic Scene Dataset. To evaluate the rendering quality of each method, we employed two widely recognized error metrics: peak signal-to-noise ratio (PSNR) and perceptual similarity (LPIPS) as defined by \cite{zhang2018unreasonable}. Additionally, due to minor differences observed in our ablation study's results, we incorporated the structural similarity index (SSIM) for a more thorough assessment.
\begin{table*}
\caption{Assessing novel view synthesis outcomes, we measure performance using the mPSNR and mSSIM metrics, benchmarked against established methods. The evaluation is conducted on the iPhone dataset \cite{gao2022monocular}.}
\label{tab:sota1}
\resizebox{\linewidth}{!}{
\begin{tabular}{lcccccccc}
\toprule
Method    & Apple         & Block         & Paper-windmill & Space-out     & Spin          & Teddy         & Wheel         & Average       \\ \hline
NSFF \cite{li2021neural}     & 17.54 / 0.750 & 16.61 / 0.639 & 17.34 / 0.378  & 17.79 / 0.622 & 18.38 / 0.585 & 13.65 / 0.557 & 13.82 / 0.458 & 15.46 / 0.569 \\
Nerfies \cite{park2021nerfies}   & 17.64 / 0.743 & 17.54 / 0.670 & 17.38 / 0.382  & 17.93 / 0.605 & 19.20 / 0.561 & 13.97 / 0.568 & 13.99 / 0.455 & 16.45 / 0.569 \\
HyperNeRF \cite{park2021hypernerf} & 16.47 / 0.754 & 14.71 / 0.606 & 14.94 / 0.272  & 17.65 / 0.636 & 17.26 / 0.540 & 12.59 / 0.537 & 14.59 / 0.511 & 16.81 / 0.550 \\
T-NeRF  \cite{gao2022monocular}  & 17.43 / 0.728 & 17.52 / 0.669 & 17.55 / 0.367  & 17.71 / 0.591 & 19.16 / 0.567 & 13.71 / 0.570 & 15.65 / 0.548 & 16.96 / 0.577 \\
RoDynRF \cite{liu2023robust}    & 18.73 / 0.722 & 18.73 / 0.634 & 16.71 / 0.321  & 18.56 / 0.594 & 17.41 / 0.484 & 14.33 / 0.536 & 15.20 / 0.449 & 17.09 / 0.534 \\
Our      & 19.53 / 0.691 & 19.74 / 0.626 & 17.66 / 0.346  & 18.11 / 0.601 & 19.79 / 0.516 & 14.51 / 0.509 & 14.48 / 0.430 & 17.69 / 0.531 \\ \bottomrule
\end{tabular}
}
\end{table*}

\begin{table}
    \centering
    \caption{Evaluation of the whole module, RBCT module, and temporal module on the Nvidia Dynamic Scene Dataset \cite{yoon2020novel} (Balloon 2 scene).}
    \label{tab:combined}
    
    \begin{minipage}{.32\linewidth}
      \centering
      \resizebox{\linewidth}{!}{
      \begin{tabular}{l|ccc}
        \toprule
                         & PSNR$\uparrow$ & SSIM$\uparrow$ & LPIPS$\downarrow$ \\ \hline
        A) w/o MV         & 26.01          & 0.8330         & 0.061             \\
        B) w/o Four-layer & 25.32          & 0.8205         & 0.070             \\
        C) w/o GSTF       & 26.73          & 0.8537         & 0.051             \\
        D) w/o RBCT       & 26.40          & 0.8434         & 0.057             \\
        E) w/o $\mathcal{L}_{small}$ & 26.86  & 0.8597     & 0.048             \\ \hline
        Full              & \textbf{26.86} & \textbf{0.8602} & \textbf{0.048}    \\  \bottomrule
      \end{tabular}
      }
      \caption*{(a) Whole Module}
    \end{minipage}%
    \begin{minipage}{.32\linewidth}
      \centering
      \resizebox{\linewidth}{!}{
      \begin{tabular}{l|ccc}
        \toprule
                  & PSNR$\uparrow$ & SSIM$\uparrow$ & LPIPS$\downarrow$ \\ \hline
        A) w/o CTT & 26.66          & 0.8491         & 0.060             \\
        B) w/o RT  & 26.16          & 0.8360         & 0.061             \\
        C) w/o GRSPE & 26.30        & 0.8410         & 0.056             \\
        D) RT to CTT & 26.69        & 0.8564         & 0.048             \\ \hline
        Full      & \textbf{26.86} & \textbf{0.8602} & \textbf{0.048}    \\  \bottomrule
      \end{tabular}
      }
      \caption*{(b) RBCT Module}
    \end{minipage}%
    \begin{minipage}{.32\linewidth}
      \centering
      \resizebox{\linewidth}{!}{
      \begin{tabular}{l|ccc}
        \toprule
               & PSNR$\uparrow$ & SSIM$\uparrow$ & LPIPS$\downarrow$ \\ \hline
        A) w/o CTT & 26.66          & 0.8491         & 0.060             \\
        B) w/o GSTF & 26.76         & \textbf{0.8625} & 0.050             \\ \hline
        Full      & \textbf{26.86} & 0.8602         & \textbf{0.048}    \\ \bottomrule
      \end{tabular}
      }
      \caption*{(c) Temporal Module}
    \end{minipage}
\end{table}
Our method demonstrates significant advancements, outperforming existing state-of-the-art techniques in five of the seven tested scenarios. This improvement is particularly evident in the average PSNR increase of 1dB and a notable 20\% reduction in LPIPS error, underscoring a substantial enhancement in perceptual quality compared to real images. Moreover, we extended our evaluation following DyCheck's methodology \cite{gao2022monocular} for the iPhone dataset, detailed in Table \Cref{tab:sota1}, where we report masked PSNR and SSIM scores. Given that a significant number of scenes within this dataset are long sequences, and considering our method's limitations in effectively modeling such sequences, notable improvements are limited. Nevertheless, our method demonstrates comparable performance to established methods and even exhibits slight enhancements in select scenarios. These outcomes serve as a compelling demonstration of the superior effectiveness of our framework in restoring intricate scene content.

\begin{figure}
  \centering
    \includegraphics[width=\linewidth]{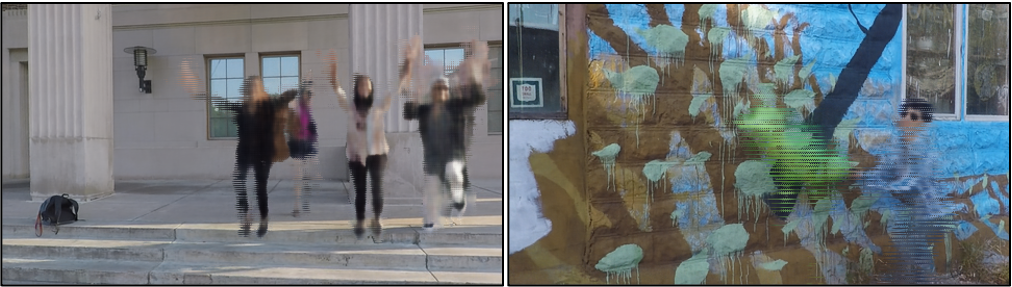}
    \caption{In the absence of global ray sampling point coordinate embedding, the synthesized view displays stripes.}
    \label{fig:6}
\end{figure}
\subsection{Ablation Study}
\label{sec:abstudy}
To validate the effectiveness of our proposed system components, we conduct an ablation study on the Dynamic Scene Dataset \cite{yoon2020novel}.
\subsubsection{Evaluate the whole module} 
\begin{wrapfigure}{r}{0.5\textwidth}
  \centering
    \includegraphics[width=\linewidth]{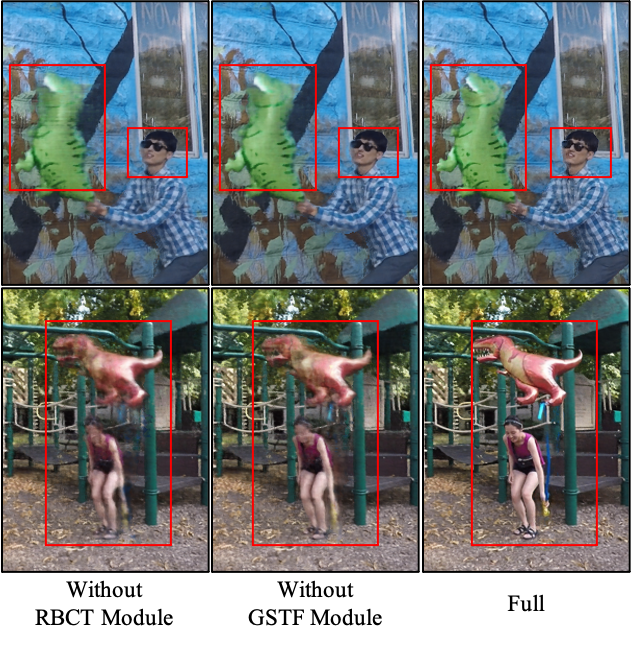}
    \caption{The qualitative results of the ablation experiments on the Nvidia Dynamic Scene Dataset \cite{yoon2020novel} (Balloon 1 and Balloon 2 scene). From right to left, showcasing unused GSTF module, unused RBCT module, and the complete model.}
    \label{fig:5}
\end{wrapfigure}
In \Cref{tab:combined}, we present a detailed comparison between our complete system and its variants, each lacking a specific module: A) multi-view aggregation, B) an additional four-layer MLP, C) Global Spatio-Temporal Filter module, D) Ray-Based Cross-Time Aggregation Module, and E) regularization scene flow loss.  As indicated in the first two rows of \Cref{tab:combined}, the absence of multi-view aggregation and the additional four-layer MLP markedly diminish the quality of view synthesis, with a decrease in PSNR by 0.75\% and 2.61\% respectively, and SSIM by 1.29\% and 2.81\% respectively. Additionally, LPIPS scores increased by 25.00\% and 27.08\%, signaling a noticeable degradation in image quality and accuracy. These two components, therefore, are critical in enhancing the fidelity and precision of the synthesized views. We observe that removing the Global Spatio-Temporal Filter module and the Ray-Based Cross-Time Aggregation Module also impacts the system's performance, though to a slightly lesser extent compared to the first two components, with PSNR decreasing by 2.08\% and 0.63\%, and SSIM by 2.23\% and 0.44\% respectively. Additionally, the removal of the regularization scene flow loss demonstrates a relatively minor impact on the quality of view synthesis, with a less pronounced decrease in performance metrics compared to the other modules. This suggests that while this component aids in fine-tuning the system, its absence does not drastically compromise the overall effectiveness.

\subsubsection{Evaluate the RBCT module} Furthermore, we investigate the impact of the internal structure of the RBCT module on the model-view synthesis performance, as summarized in \Cref{tab:combined}. Specifically, we examine the effects of the following variations: A) without using Cross-time Transformer, B) without using Ray Transformer, C) without using global ray sampling point coordinate embedding, and D) using Ray Transformer first, followed by Cross-time Transformer. Excluding the Cross-time Transformer led to a moderate decline in synthesis quality, as indicated by a 0.74\% decrease in PSNR and a 1.29\% drop in SSIM, coupled with a notable 25\% increase in LPIPS. The omission of the Ray Transformer had a more pronounced impact on performance, with a 2.61\% reduction in PSNR, a 2.82\% decrease in SSIM, and a 27.08\% rise in LPIPS. This highlights the Ray Transformer's critical role in maintaining high-quality synthesis. Furthermore, removing the global ray sampling point coordinate embedding also negatively affected the results, leading to a 2.08\% reduction in PSNR, a 2.23\% decrease in SSIM, and a 16.67\% increase in LPIPS. Sequentially applying the Ray Transformer and the Cross-time Transformer slightly improved some metrics compared to using the full module configuration, with a 0.11\% increase in PSNR and a 0.53\% rise in SSIM, while maintaining stable LPIPS. Therefore, our experiments demonstrated that the absence of key components, especially the Ray Transformer and the global ray sampling point coordinate embedding, significantly compromises view synthesis quality. In \Cref{fig:6}, the absence of global ray sampling point coordinate embedding resulted in stripes in the dynamic synthesis region. Although the other two experiments have a minimal impact on the model, the differences are discernible.

\subsubsection{Evaluate the temporal module}
\begin{wraptable}{r}{0.5\textwidth}
    \centering
    \resizebox{\linewidth}{!}{
    \begin{tabular}{l|ccc}
    \hline
        & PSNR$\uparrow$ & SSIM$\uparrow$ & LPIPS$\downarrow$ \\ 
    \hline
    w/ GSTF & \textbf{21.36} & \textbf{0.6597} & \textbf{0.260} \\
    w/o GSTF & 20.76 & 0.6129 & 0.350 \\ 
    \hline
    \end{tabular}
    }
    \caption{Evaluation of GSTF module for Dynamic region on the Nvidia Dynamic Scene Dataset \cite{yoon2020novel} (Balloon 2 scene).}
    \label{tab:GSTF_dyn}
\end{wraptable}
To demonstrate the effectiveness of our proposed frequency-domain timing module, as summarized in \Cref{tab:combined}. Specifically, A) without using Cross-time Transformer, B) Global Spatio-Temporal Filter module. The model without CTT shows a 1.11\% improvement in PSNR, a 0.41\% improvement in SSIM, and a 33.33\% reduction in LPIPS compared to the full model. Conversely, without GSTF, although the PSNR improves by 0.1\%, SSIM increases by 2.34\%, and LPIPS decreases by 16.67\%. In the Full model, we observe more substantial improvements. Compared to A), the Full model shows a 1.2\% increase in PSNR, a 1.11\% improvement in SSIM, and a 20.00\% reduction in LPIPS. In comparison to B), the Full model exhibits a 0.1\% increase in PSNR, a 0.23\% decrease in SSIM, and a 4.00\% reduction in LPIPS. This indicates that simultaneous utilization of CTT and GSTF significantly enhances the quality of the novel view, with a more pronounced improvement in perceptual quality. From the perspective of the error metric, there may not be a significant disparity between the two approaches, but utilizing them simultaneously can be complementary and enhance quality of novel view. 

\subsubsection{Evaluate the GSTF module}
\begin{wrapfigure}{r}{0.5\textwidth}
  \centering
    \includegraphics[width=\linewidth]{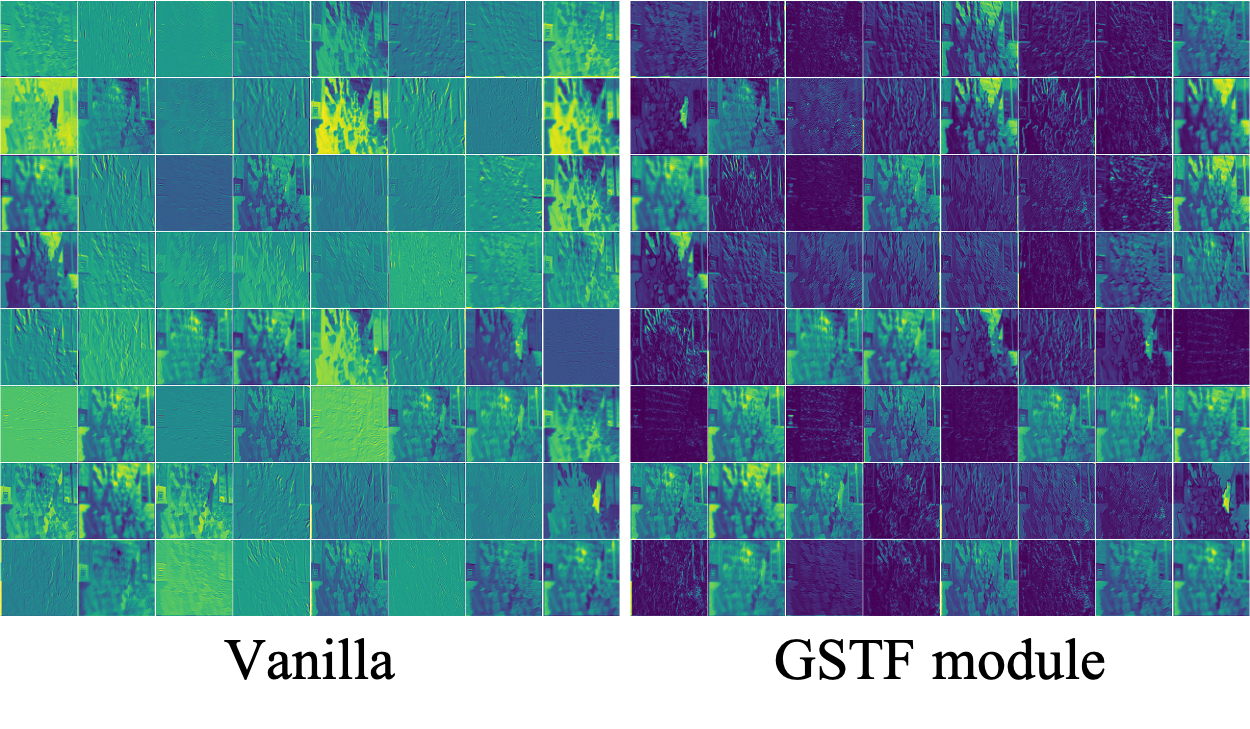}
    \caption{The left is the vanilla feature maps extracted from Resnet34 \cite{He_2016_CVPR}. In contrast, the right displays a feature map refined using our GSTF module. This comparison clearly demonstrates that our GSTF module enhances the extraction of high-frequency details, such as texture and contour, while effectively filtering out low-frequency information.}
    \label{fig:GSTF_ab}
\end{wrapfigure}
In \Cref{fig:GSTF_ab}, we present a visual comparison between baseline features extracted from Resnet34 and the features refined by our GSTF module. Our GSTF is designed with the specific goal of capturing detailed contours and high-frequency texture information, ensuring the preservation of sharp textures in the reconstructed view. A quantitative evaluation in \Cref{tab:GSTF_dyn} further underscores the impact of the GSTF module, particularly in the context of dynamic scenes. The results reveal a substantial 2.1\% increase in PSNR, a noteworthy 4.68\% improvement in SSIM, and a significant 25.38\% reduction in LPIPS when utilizing the GSTF module (\textit{w/ GSTF}) compared to the configuration without GSTF (\textit{w/o GSTF}). In \Cref{tab:combined}, the GSTF module makes a marginal contribution to the overall improvement. This is primarily because our GSTF is applied to dynamic areas, which represent just one-fifth of the total in the Balloon 2 evaluation scene.

\subsubsection{Ablation study qualitative results} 
\begin{wrapfigure}{r}{0.5\textwidth}
    \centering
    \includegraphics[width=\linewidth]{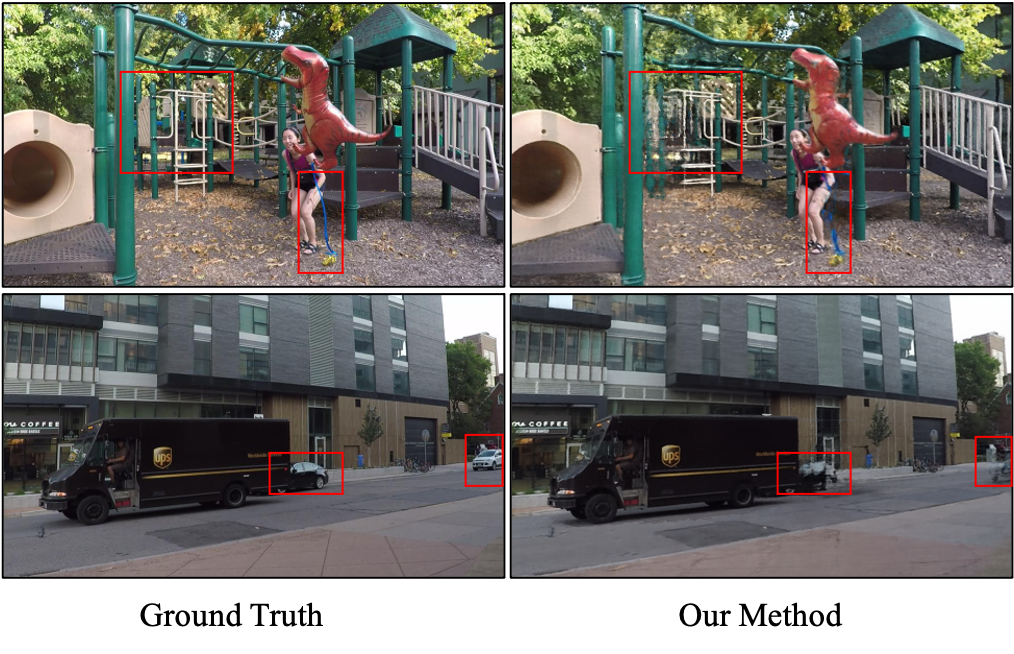}
    \caption{Limitations of our model, take the Nvidia Dynamic Scene Dataset \cite{yoon2020novel} (Balloon 1 and truck scene) as an example.}
    \label{fig:7}
\end{wrapfigure}
The \Cref{fig:5} illustrates our primary contributions, the Global Spatio-Temporal Filter (GSTF) and Ray-Based Cross-Time (RBCT) modules. These modules play pivotal roles in enhancing the quality of synthesized views. In the absence of the RBCT module, the resulting synthesized view lacks intricate surface details and may exhibit noticeable artifacts. Conversely, in the absence of the GSTF module, the synthesized view experiences a loss of edge information, resulting in a perceptible blurring effect.

\subsection{Limitations}
As shown in \Cref{fig:7}, compare with the ground truth, we can observe that in the playground scene, the rendering of the railings and balloon ropes appears to be blurred. This is because we are using multi-frame aggregation of feature vectors, which enables us to aggregate more information but also results in some details that may be discarded or merged into other features, making it impossible to fully express the original data. Moreover, due to the small non-rigid deformations of these parts, our method cannot handle them well, resulting in blurring when rendering new views. In the truck scene, we only input the adjacent 2 frames of the current frame. Therefore, when the time is changed for rendering after the $1st$ frame with a fixed sequence number, the synthesis effect of the long sequence of unseen frames is not optimal. For example, the synthesis result of the $11th$ frame in the figure shows that the hidden car behind the truck is very blurred due to the lack of feature vector information provided by adjacent frames.

\section{Conclusion}
In this work, we aim to introduce a novel dynamic neural render field framework for dynamic monocular videos, which enables high-quality rendering of novel views. To achieve this goal, we extend recent ideas in multi-view aggregation to time-varying NeRF, enabling the modeling of complex motion. Specifically, we introduce RBCT and GSTF modules to model motion from the time domain and frequency domain, respectively. Our experimental results show that these proposed modules significantly improve the performance of time-varying NeRF with multi-view aggregation when rendering new views. While our work represents a promising exploration of time-varying NeRF for multi-view aggregation, there are still some limitations. It is worth noting that our current method may not perform well when rendering novel views of long sequences of videos. One potential solution to improve performance is to increase the length of the aggregate view, but this approach requires significant computing resources. Fortunately, recent developments such as TensoRF and 3D Gaussian splatting offer potential solutions to these challenges.

\section*{Acknowledgments}
This work is supported by the UK Medical Research Council (MRC) Innovation Fellowship under Grant MR/S003916/2, International Exchanges 2022 IEC$\backslash$NSFC$\backslash$223523 and Securing the Energy/Transport Interface EP/X037401/1.

\bibliographystyle{elsarticle-num-names.bst} 
\bibliography{main}

\end{document}